# Decentralized Federated Learning via Mutual Knowledge Transfer

Chengxi Li, Gang Li, *Senior Member, IEEE,* Pramod K. Varshney, *Life Fellow, IEEE*

**Abstract**—**In this paper, we investigate the problem of decentralized federated learning (DFL) in Internet of things (IoT) systems, where a number of IoT clients train models collectively for a common task without sharing their private training data in the absence of a central server. Most of the existing DFL schemes are composed of two alternating steps, i.e., model updating and model averaging. However, averaging model parameters directly to fuse different models at the local clients suffers from client-drift especially when the training data are heterogeneous across different clients. This leads to slow convergence and degraded learning performance. As a possible solution, we propose the decentralized federated learning via mutual knowledge transfer (Def-KT) algorithm where local clients fuse models by transferring their learnt knowledge to each other. Our experiments on the MNIST, Fashion-MNIST, CIFAR-10, and CIFAR-100 datasets reveal that the proposed Def-KT algorithm significantly outperforms the baseline DFL methods with model averaging, i.e., Combo and FullAvg, especially when the training data are not independent and identically distributed (non-IID) across different clients.**

## I. INTRODUCTION

NOWADAYS, an unprecedented amount of data are being generated by devices such as smart phones in the booming applications of Internet of things (IoT) including smart city and smart factory [39]-[42], which has facilitated the emergence of data-driven methods such as machine learning (ML) [48], [50]. In the traditional ML paradigm, models are trained with large datasets collected by a central server. However, in many practical applications of IoT, data may be privacy-sensitive and it may be costly to aggregate large datasets at a central server. In such cases, the training data are often distributed on different IoT clients such as sensors [47], phones or other information sources [3] where they are generated. To train models using decentralized training data in IoT systems, federated learning (FL) is emerging as a new framework for learning tasks while not requiring the clients to transmit their raw datasets, thereby reducing the communication cost and at the same time guaranteeing data privacy [1]-[3], [20]-[23],[49].

In a general FL system, a central server coordinates the training task based on the data at the clients [4], [26]. As shown in Fig. 1, during each round, each participating client individually updates the model based on its local dataset and then transmits the updated model to the server. Upon receiving the models from all the participating clients during the current round, the server conducts model averaging and broadcasts the updated model to the participating clients of the next round [2]. Although the FL systems mentioned above are promising, they are confronted with many challenges. While large organizations could play the role of a central server in some IoT applications, in many FL environments, it is difficult to find a central server that is both reliable and powerful [3], [37]. Besides, a malfunction in the central server induces a single point of failure of the whole network. To overcome the above shortcomings of general FL systems that use a central server, decentralized FL (DFL) methods which do not need a central server are well worth investigating. In DFL, the clients exchange their model parameters directly in a peer-to-peer manner [4]. Actually, the advantages of DFL lie in not only eliminating the single point failure of the central server, but also attaining scalability in an inexpensive manner since no additional infrastructure is necessary [3].

Gossip averaging is a well known method in a variety of decentralized algorithms [5], [27], [28], where different clients in the network exchange information in a peer-to-peer manner without the help of a central server. In [6], [8] and [9], gossip averaging has been employed with stochastic gradient descent (SGD) to train deep learning models in a decentralized way, which manifests excellent convergence properties. Built on the above prior work, gossip averaging was later applied to DFL schemes, where model updating and model averaging are implemented in an alternating manner at the local clients [7],[10],[11],[36]. While in [7],[10] and [11] the clients send and average full sets of model parameters, the Combo algorithm in [36] has been proposed by letting the clients transmit and average model segments in order to make better use of the communication resources without impacting the convergence rate. Although the aforementioned DFL algorithms do overcome some of the difficulties confronting the general FL systems that require a central server, they all employ model averaging to fuse models at the local clients, which is not always very efficient due to the following reasons. During the training process, the models are updated locally

This work was supported in part by National Natural Science Foundation of China under Grants 61790551 and 61925106. Corresponding author: Gang Li. Email: gangli@mail.tsinghua.edu.cn.

C. Li and G. Li are with the Department of Electronic Engineering, Tsinghua University, Beijing, 100084, China.

P. K. Varshney is with the Department of Electrical Engineering and Computer Science, Syracuse University, Syracuse, NY 13244, USA.

towards the local optimum of each client's loss function, and averaging the model parameters from different clients results in a model moving towards the averaged result of the corresponding local optima. Since different clients own distinct sets of training data and those datasets normally have no overlap or even have different distributions, which is known as data heterogeneity [12], [29]-[31], the optimum of each client's loss function may be quite far from each other, which is also far away from the global optimum. Furthermore, with heterogeneous training data, the difference between the averaged result of the local optima and the global optimum naturally arise, which indicates that the averaged model may be quite different from the global optimum. This phenomenon is known as client-drift [12]. From this perspective, averaging two models that have been trained on heterogeneous data results in slow and unstable convergence [12] under the DFL scheme. This may happen even though all the clients are initialized identically. With degraded learning performance in the presence of data heterogeneity, model averaging of the DFL algorithms in [7],[10],[11],[36] does not make full use of the training datasets to enhance the generalization ability of the trained networks. In fact, instead of suffering from performance degradation when learning from heterogeneous and decentralized data, one should be able to exploit data heterogeneity to improve the generalization ability of the trained networks. However, it remains an open problem to overcome the shortcomings of the baseline DFL methods that use model averaging [7],[10],[11],[36] and to find new methods which could avoid severe degradation of the learning performance caused by data heterogeneity.

In order to prevent the client-drift from impacting the convergence rate and the training stability and to take full advantage of the heterogeneous data, several FL algorithms have been proposed in the literature [12],[16],[17],[53],[54]. For example, Li et al. made some modifications to FedAvg [2] by incorporating a proximal term in the original objective so as to mitigate the adverse impact of data heterogeneity on the stability of the convergence behavior [17]. In [53], an FL algorithm based on normalized averaging was proposed in order to eliminate the inconsistency of the mismatched objectives and to maintain a fast convergence rate. In [54], a snapshotting scheme was employed under the FL framework, where the updating of the model parameters and the updating of the mixing parameter are decoupled. However, all of these methods have been proposed for general FL scenarios assuming the availability of a central server, and they cannot be applied to DFL schemes directly.

For numerous problems, neural networks that attain excellent performance have been designed. They, however, always contain an exceedingly large number of model parameters, which restricts their utilization in platforms with limited memory and in applications requiring fast execution. In order to develop more compact models that behave as well as the large ones, distillation-based methods have been proposed by letting the small-sized models imitate the soft outputs of the larger ones [13],[18],[19],[43]-[46]. Among the distillation methods, deep mutual learning (DML) [13] adopts two student networks that learn on a common dataset collaboratively and simultaneously by teaching each other during the process of training. It is shown that, each student model attains better learning performance compared with that achieved when each model is trained separately as done in the conventional way. The success of DML is achieved because the two student networks are initialized differently, which enables them to learn distinct knowledge from the common data samples and to transfer their knowledge to each other during the training process. However, to the best of our knowledge, the advantages of the mutual knowledge transfer (MKT) strategy adopted by DML have not been exploited to enhance the learning performance of DFL tasks.

In this paper, to perform DFL tasks in IoT systems in the presence of data heterogeneity, we propose a new algorithm called decentralized federated learning via mutual knowledge transfer (Def-KT), which effectively incorporates the advantages of MKT into the DFL framework to avoid the negative impact of client-drift. In each round of the proposed Def-KT algorithm, two steps are implemented sequentially, i.e., model updating and model fusion. In the first step, we randomly choose[1] a fixed number of clients, each of which updates the local model by performing a number of training passes over its private dataset via stochastic gradient descent (SGD) and then sends the fine-tuned model to another randomly picked client. In the second step, inspired by DML [13], each client that has received a model in the current round fuses its local model and the received one using MKT rather than averaging them as done in [7],[10],[11] and [36]. The motivation of doing this is explained as follows. In the DFL methods with model averaging, the original incentive of averaging local models that have been trained on different training datasets is to obtain a model that performs well on data samples drawn from all of the datasets. In other words, the averaged model is supposed to acquire knowledge on different datasets. However, it is quite difficult to achieve the above goal with heterogeneous training data, negatively impacted by client-drift. In contrast, MKT enables two models with different knowledge to learn from each other and the resulting models obtain knowledge indirectly from both models. Based on the above analysis, it is intuitive that MKT can be adopted as a better alternative for model fusion under the DFL scheme. By doing this, the knowledge previously learnt by the two models at two different clients can be retained in the resulting model of MKT, with both heterogeneous training data and homogeneous data, which guarantees a better generalization ability on new data samples. To demonstrate the superiority of the proposed Def-KT algorithm, we run experiments on the MNIST [14], Fashion-MNIST [25], CIFAR-10 [15], and CIFAR-100 [51] datasets for image classification tasks, where Def-KT is compared with the baseline DFL methods that use model averaging, including FullAvg which performs averaging over full sets of model parameters [7],[10],[11] and Combo which performs averaging over model segments [36]. The experimental results show that the proposed method significantly outperforms FullAvg and Combo in various settings, observing that the former converges faster and attains more stable learning performance than the latter. The main

---

[1] The clients participating in the first step are randomly chosen by us to simulate the practical case where clients do not always volunteer to participate in the training process unless they are plugged-in or charged, as elaborated in [2].

contributions of our paper are listed as follows:

1) To avoid performance degradation induced by client-drift, we innovatively incorporate the advantages of MKT into the DFL schemes in the presence of data heterogeneity. Although MKT adopted by the proposed Def-KT algorithm is inspired by DML [13], its application to DFL problems is not old wine in a new bottle, since the rationale behind DML and MKT in Def-KT are very different from each other. To be more specific, the student models in DML are initialized differently, which enables them to learn different knowledge on the same data samples. In contrast, in the model fusion stage of Def-KT, the two models, although initialized to be the same, are trained on non-overlapping datasets and thus have distinct "expertise". This enables the two models to transfer their knowledge to each other so that a model with better generalization ability can be obtained.

2) We run experiments on four popular datasets for image classification tasks using the proposed Def-FL algorithm as well as the baseline methods with model averaging, namely FullAvg and Combo. Our experimental results demonstrate the superiority of Def-KT over the baseline methods.

The rest of this paper is organized as follows. The considered problem is formulated in Section II. The proposed Def-KT algorithm and its rationale are presented in Section III. In Section IV, we provide experimental results to demonstrate the superiority of the proposed algorithm over the baseline methods and discuss the results. Finally, the conclusion of this paper is given in Section V.

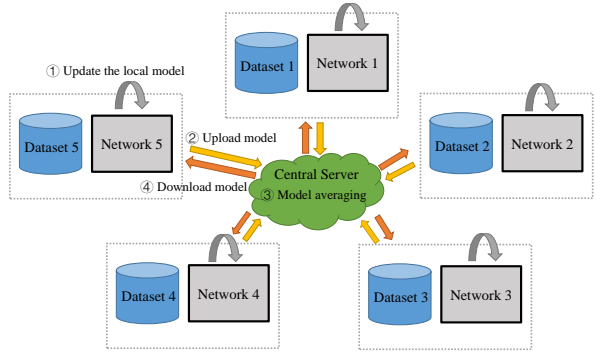

Fig. 1. A general federated learning framework with a central server in IoT systems.

## II. PROBLEM FORMULATION

The considered DFL problem is formulated as follows. Suppose there are $K$ IoT clients in the network. Each client possesses a labeled image dataset $\mathcal{D}_k := \left\{ \left( x_i^k, y_i^k \right) \right\}_{i=1}^{N_k}$, $k = 1, ..., K$, where $x_i^k$ is the $i$-th data sample of the $k$-th client, $y_i^k \in \{1, 2, ..., C\}$ is the corresponding label among $C$ classes, and $N_k$ denotes the number of training samples owned by the $k$-th client. The datasets of different clients may be drawn from different distributions $P_k$, $k = 1, ..., K$, mainly due to the fact that those training data have been generated and collected by different clients in a non-identical manner. The goal is to train models for the image classification task with peer-to-peer communications among the clients in the absence of a central server. To deal with the privacy concerns, it is the model parameters that are communicated in the network instead of raw datasets which are deemed privacy-sensitive. Before training starts, each client is initialized with a model of the same architecture as well as the same parameters.

In the DFL problem considered in this paper, the following assumptions are made, which comply with the resource-constrained learning environments. 1) Only a small subset of clients participate in each round of the training stage, which are referred to as the participating clients. 2) In each round, only a fraction of the participating clients train their local models on private datasets and those clients transmit the fine-tuned models to another set of clients. The reason of making the above assumptions is three-fold. First, each client only volunteers to take part in the training under certain circumstances such as when the device is plugged-in [2], [29]. Second, the computational resources among the participating clients may not be balanced and it may be impractical to let them perform the same computational tasks during each round [38]. Third, the communication cost is one of the dominating factors and it is unrealistic to allow all the clients to transmit their models to each other simultaneously in a single round [2].

## III. DECENTRALIZED FEDERATED LEARNING VIA MUTUAL KNOWLEDGE TRANSFER (DEF-KT)

### A. The Proposed Method

In this paper, a new algorithm named Def-KT is proposed to tackle the problem formulated in Section II. It is presented as Algorithm 1. The glossary of important notations corresponding to different variables used in the proposed method is provided in Table I. At the beginning of the $t$-th round, $Q$ clients $(2Q \ll K)$ are randomly selected, whose indices are denoted by $\mathcal{I}_t^A = \left\{ k_t^1, k_t^2, ..., k_t^Q \right\}$. Then, each of the $Q$ clients makes $M$ training passes over its private dataset $\mathcal{D}_{k_t^j}$ to update its local model[2] $\mathbf{w}_t^{k_t^j}$ in parallel via stochastic gradient descent (SGD) with local minibatch size $B_1$ and learning rate $\eta_0$, which is shown as follows

$$\tilde{\mathbf{w}}_t^{k_t^j} \leftarrow \text{SGD}_{B_1, M} \left( \mathbf{w}_t^{k_t^j}, \mathcal{D}_{k_t^j} \right), \forall j = 1, ..., Q. \quad (1)$$

After that, another $Q$ clients are randomly selected to communicate with the clients in set $\mathcal{I}_t^A$ in a peer-to-peer manner, whose indices are denoted by $\mathcal{I}_t^B = \left\{ k_t^{Q+1}, k_t^{Q+2}, ..., k_t^{2Q} \right\}$. In this paper, it is assumed that set $\mathcal{I}_t^A$ and set $\mathcal{I}_t^B$ have no overlap. In other words, a participating client in a round would either update and send its model or receive a model from the other clients. It does not do both[3]. To be more specific, the $k_t^j$-th client transmits its updated model $\tilde{\mathbf{w}}_t^{k_t^j}$ to the $k_t^{j+Q}$-th client, $j = 1,...,Q$. In this case, $2Q$ clients participate in each round and only half of them train their local models on their private datasets. The overall communication overhead in each round comes from the transmission of $Q$ sets of model parameters in the network. By assuming $Q$ to be a small number, only a small subset of clients undertake heavy computational tasks and the communication overhead is affordable for a system under resource constraints. This ensures that the proposed algorithm can be easily implemented under practical scenarios. It is also worth pointing out that, although discussed under the above setting, the proposed algorithm could be generalized to other cases where the number of clients that can afford to perform local training tasks and the communication overhead in the network are different from those in this paper. The analysis and experiments under more generalized settings are not included in this paper due to page limitations.

---

[2] In this paper, we assume that the models at different clients share the same architecture. For a concise representation, a model is unambiguously represented by a vector containing all the parameters. For example, "its local model $\mathbf{w}_t^{k_t^j}$" means "its local model with parameters contained in a vector $\mathbf{w}_t^{k_t^j}$".

TABLE I
GLOSSARY OF NOTATIONS IN DEF-KT

| Symbol | Descriptions |
|---|---|
| $K$ | The number of IoT clients in the network. |
| $\mathcal{D}_k$ | The private dataset of the $k$-th client, $k = 1,...,K$. |
| $Q$ | The number of transmitting clients in each round. |
| $T$ | The total number of rounds. |
| $\mathbf{w}_0$ | The vector containing all the initial parameters of the model. |
| $B_1$ | The local minibatch size of local updating. |
| $M$ | The number of training passes of local updating. |
| $\eta_0$ | The learning rate of local updating. |
| $B_2$ | The local minibatch size of MKT. |
| $E$ | The number of training passes of MKT. |
| $(\eta_1, \eta_2)$ | The learning rates of MKT. |

Upon receiving the model from the $k_t^j$-th client, the $k_t^{j+Q}$-th client mixes the knowledge of the received model $\tilde{\mathbf{w}}_t^{k_t^j}$ and its local model $\mathbf{w}_t^{k_t^{j+Q}}$ by letting the two models transfer knowledge to each other, which is motivated by DML. Next, the MKT strategy in the proposed Def-KT algorithm will be elaborated in more detail. First, $\mathcal{D}_{k_t^{j+Q}}$ is split into $L$ mini-batches of size $B_2$ denoted by $\{\mathcal{B}_l, l = 1,...L\}$ to be fed in models $\tilde{\mathbf{w}}_t^{k_t^j}$ and $\mathbf{w}_t^{k_t^{j+Q}}$. It is worth noting that, distinct "expertise" of the two models results in different outputs of their softmax layer, a.k.a. soft predictions, on the same mini-batch of data samples. Based on that, to transfer their knowledge to each other, model $\tilde{\mathbf{w}}_t^{k_t^j}$ and model $\mathbf{w}_t^{k_t^{j+Q}}$ try to

imitate the output of the other one by minimizing the loss functions given as

$$\text{Loss}_1\left(\tilde{\mathbf{w}}_t^{k_t^j}, \mathcal{B}_l, \mathcal{P}_{2,l}\right) = L_C\left(\mathcal{P}_{1,l}, \mathcal{Y}_l\right) + D_{KL}\left(\mathcal{P}_{2,l} \| \mathcal{P}_{1,l}\right), \quad (2)$$

and

$$\text{Loss}_2\left(\mathbf{w}_t^{k_t^{j+Q}}, \mathcal{B}_l, \mathcal{P}_{1,l}\right) = L_C\left(\mathcal{P}_{2,l}, \mathcal{Y}_l\right) + D_{KL}\left(\mathcal{P}_{1,l} \| \mathcal{P}_{2,l}\right), \quad (3)$$

respectively, where $\mathcal{Y}_l = \left\{y_z^l\right\}_{z=1}^{B_2}$ denotes the set of true labels of the data samples in $\mathcal{B}_l$ with $y_z^l \in \{1,2,...,C\}$ being the true label of the $z$-th sample in $\mathcal{B}_l$. In (2) and (3), $\mathcal{P}_{1,l}$ and $\mathcal{P}_{2,l}$ are the soft predictions produced as

$$\begin{aligned}
\mathcal{P}_{1,l} &= \left\{\mathbf{p}_{1,l,z}\right\}_{z=1}^{B_2} = \text{model}\left(\mathcal{B}_l, \tilde{\mathbf{w}}_t^{k_t^j}\right), \forall l, \\
\mathcal{P}_{2,l} &= \left\{\mathbf{p}_{2,l,z}\right\}_{z=1}^{B_2} = \text{model}\left(\mathcal{B}_l, \mathbf{w}_t^{k_t^{j+Q}}\right), \forall l,
\end{aligned} \quad (4)$$

where $\mathbf{p}_{1,l,z} = \left[p_{1,l,z}^1, p_{1,l,z}^2,....,p_{1,l,z}^C\right]$ and $\mathbf{p}_{2,l,z} = \left[p_{2,l,z}^1, p_{2,l,z}^2,....,p_{2,l,z}^C\right]$ are two vectors of length $C$ denoting the soft predictions of the $z$-th data sample in the mini-batch $\mathcal{B}_l$. $L_C(\bullet)$ in (2) and (3) denotes the cross entropy error between the true labels and the soft predictions, which is given by

$$\begin{aligned}
L_C\left(\mathcal{P}_{1,l}, \mathcal{Y}_l\right) &= -\sum_{z=1}^{B_2} \mathbf{h}^T\left(y_z^l\right) \log \mathbf{p}_{1,l,z}, \\
L_C\left(\mathcal{P}_{2,l}, \mathcal{Y}_l\right) &= -\sum_{z=1}^{B_2} \mathbf{h}^T\left(y_z^l\right) \log \mathbf{p}_{2,l,z},
\end{aligned} \quad (5)$$

where $\mathbf{h}(y)$ is the $C \times 1$ one-hot vector and all the elements in $\mathbf{h}(y)$ are zero except that the $y$-th element is one. The second term in (2) and (3) is the Kullback Leibler (KL) Divergence that quantifies the match of the soft predictions of the two networks, which is given as

$$\begin{aligned}
D_{KL}\left(\mathcal{P}_{2,l} \| \mathcal{P}_{1,l}\right) &= \sum_{z=1}^{B_2} \sum_{c=1}^{C} p_{2,l,z}^c \log \frac{p_{2,l,z}^c}{p_{1,l,z}^c}, \\
D_{KL}\left(\mathcal{P}_{1,l} \| \mathcal{P}_{2,l}\right) &= \sum_{z=1}^{B_2} \sum_{c=1}^{C} p_{1,l,z}^c \log \frac{p_{1,l,z}^c}{p_{2,l,z}^c},
\end{aligned} \quad (6)$$

After updating $\tilde{\mathbf{w}}_t^{k_t^j}$ and $\mathbf{w}_t^{k_t^{j+Q}}$ by minimizing the loss functions in (2) and (3) through $E$ passes over all the mini-batches of dataset $\mathcal{D}_{k_t^{j+Q}}$ as shown in (7) and (8) in Algorithm 1, the $k_t^{j+Q}$-th client stores the resulting model $\tilde{\mathbf{w}}_t^{k_t^j}$ as its local model, which incorporates the knowledge of both models participating in the MKT process. To make it clear, Fig. 2 presents the schematic of a round of the proposed Def-KT algorithm.

### B. The Rationale Behind the Proposed Method

Different from the model averaging strategy adopted by the prior DFL methods [7],[10],[11],[36], the proposed method utilizes MKT to fuse models at the local clients, which effectively avoids the negative impact of client-drift in the local updates. Next, we explicitly explain the motivation of using MKT under the DFL framework. For brevity, the subsequent notations follow those in Section III.A.

---

[3] Actually, the proposed method can be easily extended to other cases where a participating client both transmits and receives models in a round. The extended models are not analyzed in this paper since the network topology is not the main focus of this paper.

---

**Algorithm 1: Def-KT (The proposed method)**

*Input:* initial parameters $\mathbf{w}_0$

*Initialization:* All clients are initialized with the same model with parameters $\mathbf{w}_0$.

**For** $t = 0,...,T$ **do**

    Randomly select a set $\mathcal{I}_t^A = \left\{ k_t^1, k_t^2, ..., k_t^Q \right\}$ of clients.

    Randomly select another set $\mathcal{I}_t^B = \left\{ k_t^{Q+1}, k_t^{Q+2}, ..., k_t^{2Q} \right\}$ of clients.

    **For** $j \in \{1,...,Q\}$ **in parallel do**

$$\tilde{\mathbf{w}}_t^{k_t^j} \leftarrow \text{SGD}_{B_1:M}\left( \mathbf{w}_t^{k_t^j}, \mathcal{D}_{k_t^j} \right).$$

        The $k_t^j$ -th client stores $\tilde{\mathbf{w}}_t^{k_t^j}$ as its local model and transmits $\tilde{\mathbf{w}}_t^{k_t^j}$ to the $k_t^{j+Q}$ -th client.

    **end**

    **For** $j \in \{1,...,Q\}$ **in parallel do**

        The $k_t^{j+Q}$ -th client does:

        $\{\mathcal{B}_l, l = 1,...L\} \leftarrow$ split $\mathcal{D}_{k_t^{j+Q}}$ into minibatches of size $B_2$

        **For** $e = 1,...,E$ **do**

            For $l \in \{1,...,L\}$ do

            Compute soft predictions $\mathcal{P}_{1,l} = \text{model}\left( \mathcal{B}_l, \tilde{\mathbf{w}}_t^{k_t^j} \right)$.

            Compute soft predictions $\mathcal{P}_{2,l} = \text{model}\left( \mathcal{B}_l, \mathbf{w}_t^{k_t^{j+Q}} \right)$.

            Update $\tilde{\mathbf{w}}_t^{k_t^j}$:

$$\tilde{\mathbf{w}}_t^{k_t^j} \leftarrow \tilde{\mathbf{w}}_t^{k_t^j} - \eta_1 \frac{\partial \text{Loss}_1\left( \tilde{\mathbf{w}}_t^{k_t^j}, \mathcal{B}_l, \mathcal{P}_{2,l} \right)}{\partial \tilde{\mathbf{w}}_t^{k_t^j}}. \quad (7)$$

            Update $\mathbf{w}_t^{k_t^{j+Q}}$:

$$\mathbf{w}_t^{k_t^{j+Q}} \leftarrow \mathbf{w}_t^{k_t^{j+Q}} - \eta_2 \frac{\partial \text{Loss}_2\left( \mathbf{w}_t^{k_t^{j+Q}}, \mathcal{B}_l, \mathcal{P}_{1,l} \right)}{\partial \mathbf{w}_t^{k_t^{j+Q}}}. \quad (8)$$

            **end**

        **end**

        Store $\tilde{\mathbf{w}}_t^{k_t^j}$ as the local model at the $k_t^{j+Q}$ -th client.

    **end**

**end**

*Until:* convergence

It is obvious that the received model $\tilde{\mathbf{w}}_t^{k_t^j}$ and the local model $\mathbf{w}_t^{k_t^{j+Q}}$ have been trained on different and possibly heterogeneous datasets, which indicates that they have distinct "expertise" on the training data. To be more specific, the received model $\tilde{\mathbf{w}}_t^{k_t^j}$ has been trained on $\mathcal{D}_{k_t^j}$ at the beginning of the $t$-th round so it performs well on $\mathcal{D}_{k_t^j}$. In contrast, the local model $\mathbf{w}_t^{k_t^{j+Q}}$ has gained much information from $\mathcal{D}_{k_t^{j+Q}}$ in some previous rounds, which makes it perform well on $\mathcal{D}_{k_t^{j+Q}}$. Since $\mathcal{D}_{k_t^j}$ and $\mathcal{D}_{k_t^{j+Q}}$ have no overlap and they are even drawn from distinct distributions, the knowledge of model $\tilde{\mathbf{w}}_t^{k_t^j}$ and model $\mathbf{w}_t^{k_t^{j+Q}}$ is different from each other. To obtain a new model that mixes the knowledge of both the received model $\tilde{\mathbf{w}}_t^{k_t^j}$ and the local model $\mathbf{w}_t^{k_t^{j+Q}}$, MKT is implemented in the proposed Def-KT algorithm. To gain more insights into the effectiveness of the proposed algorithm, the rationale behind it is discussed in more detail as follows:

1) **Learning from unseen data samples indirectly.** Before starting MKT in the $t$-th round, clearly model $\tilde{\mathbf{w}}_t^{k_t^j}$ and model $\mathbf{w}_t^{k_t^{j+Q}}$ possess distinct knowledge gained from two different training datasets, which are owned by client $k_t^j$ and client $k_t^{j+Q}$, respectively. Then, in the process of MKT at client $k_t^{j+Q}$, by imitating the output of model $\tilde{\mathbf{w}}_t^{k_t^j}$, model $\mathbf{w}_t^{k_t^{j+Q}}$ could understand and gain the knowledge possessed by model $\tilde{\mathbf{w}}_t^{k_t^j}$ with respect to the private dataset $\mathcal{D}_{k_t^j}$ of client $k_t^j$ indirectly, even though the raw dataset $\mathcal{D}_{k_t^j}$ is not available to client $k_t^{j+Q}$. This ensures that the knowledge about dataset $\mathcal{D}_{k_t^j}$ is preserved in both models participating in the MKT process despite the inaccessibility of dataset $\mathcal{D}_{k_t^j}$.

2) **Enhancing the generalization ability by synthesizing the knowledge from models with distinct expertise.** It is obvious that model $\tilde{\mathbf{w}}_t^{k_t^j}$ and model $\mathbf{w}_t^{k_t^{j+Q}}$ are trained on different datasets before MKT, which implies that the two models have developed distinct knowledge about the data. Based on that, when inputting the same data samples, the two models learn quite different representations and soft predictions. By imitating the output of the other network, the posterior entropy of the output of each network is increased, which naturally contributes to better generalization to new test data for each of the two networks.

3) **Overcoming catastrophic forgetting.** Catastrophic forgetting is a term used in continual learning, which refers to the phenomenon that a neural network loses the knowledge regarding the previously learnt tasks when it incorporates some new knowledge about the current task [32]-[34]. A specific scenario where catastrophic forgetting occurs is training a network on multiple tasks sequentially with data from different tasks presented sequentially as well. Similar to this case, during the $t$-th round, model $\tilde{\mathbf{w}}_t^{k_t^j}$ is first trained on dataset $\mathcal{D}_{k_t^j}$. After that, model $\tilde{\mathbf{w}}_t^{k_t^j}$ is sent to the $k_t^{j+Q}$ -th client where dataset $\mathcal{D}_{k_t^j}$ is no longer accessible, $j = 1,...,Q$. During the subsequent MKT process, by letting the model $\mathbf{w}_t^{k_t^{j+Q}}$ teach and learn from model $\tilde{\mathbf{w}}_t^{k_t^j}$ simultaneously, $\tilde{\mathbf{w}}_t^{k_t^j}$ obtains new capabilities based on the "new" dataset $\mathcal{D}_{k_t^{j+Q}}$ and in the meantime maintains its performance on dataset $\mathcal{D}_{k_t^j}$ since the knowledge of model $\tilde{\mathbf{w}}_t^{k_t^j}$ is not overwritten but transferred to the model $\mathbf{w}_t^{k_t^{j+Q}}$. This effectively prevents the resulting model from forgetting the information contained in dataset $\mathcal{D}_{k_t^j}$ that is no longer available.

4) **Avoiding homogenization of different models.** It has been pointed out that an obstacle for collaborative learning including DML is the tendency of homogenization of the student networks, which results from the fact that all students learn from the same entire dataset all along. The phenomenon of homogenization leads to degraded generalization ability and impacts the performance gain brought by collaboration of multiple networks [35]. In contrast, MKT in Def-KT is not affected by model homogenization. This is because the local training process at the beginning of each round at the $k_t^j$-th client, $k_t^j \in \left\{ k_t^1, k_t^2, ..., k_t^Q \right\}$, enhances the diversity of different local models and thus prevents model homogenization. In other words, the two models participating in the MKT process have been trained on totally different datasets owned by two clients and they tend to learn different feature representations on a common data sample in spite of the same initialization. From that point of view, data heterogeneity across different clients benefits the training task in the proposed Def-KT algorithm since it alleviates model homogenization and enhances the generalization ability of local models.

In the proposed Def-KT algorithm, the strength of MKT is incorporated into the DFL framework to fuse local models and produce models with better generalization ability. It is worth emphasizing that, although MKT in Def-KT is motivated by DML [13], the rationale behind them are very different from each other, as shown in Fig. 3. As introduced in Section I, the performance gain of DML is a consequence of different initializations of the student networks. In contrast, under the DFL schemes, the models of different clients are all initialized to share the same architecture and the same set of parameters. Due to the data heterogeneity across different clients, models at different clients are able to gain different knowledge on the training datasets. Based on that, transferring knowledge between different clients improves the generalization ability of the resulting models and thus contributes to faster convergence and better training stability.

## IV. EXPERIMENTS

In this section, a series of experiments are performed to evaluate the performance of the proposed algorithm on image classification tasks under the DFL schemes. Comparison with the baseline methods with model averaging is conducted to demonstrate the superiority of our method.

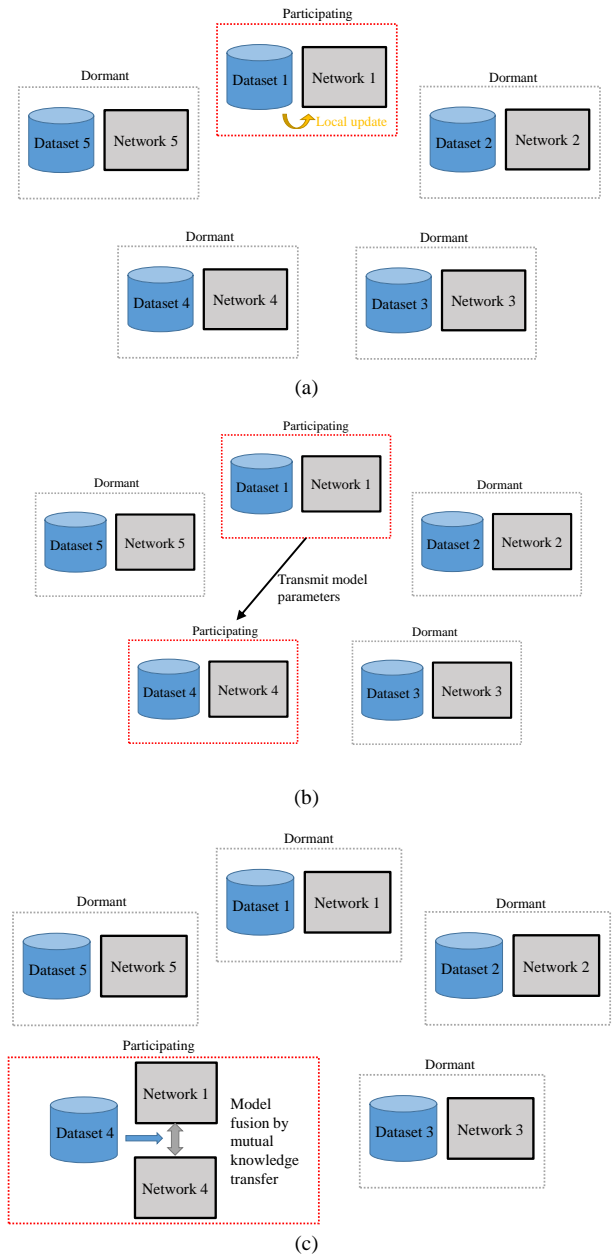

Fig. 2. Schematic of a single round of the proposed Def-KT method ($Q = 1$), where three steps are implemented sequentially. (a) Step 1: client 1 is randomly chosen to train and update its local model based on its own dataset. (b) Step 2: Client 1 transmits the updated model to another randomly chosen client 4. (c) Client 4 performs model fusion by means of mutual knowledge transfer and stores the resulting model thereof to replace its local model.

### A. Datasets

Four public datasets are used, which are MNIST [14], Fashion-MNIST [25], CIFAR-10 [15], and CIFAR-100 [51].

**MNIST** consists of 70k $28 \times 28$ images of handwritten digits of number 0 to 9, which are divided into a training set of 60k examples and a test set of 10k examples.

**Fashion-MNIST** consists of 70k $28 \times 28$ images of fashion items from 10 classes. On the whole, there are 60k images in the training set and 10k images in the test set.

**CIFAR-10** contains $32 \times 32$ RGB images of objects from 10 classes and it is originally split into a training set of 50k examples and a test set of 10k examples.

**CIFAR-100** contains $32 \times 32$ RGB images of objects from 100 classes, which is originally split into a training set of 50k examples and a test set of 10k examples.

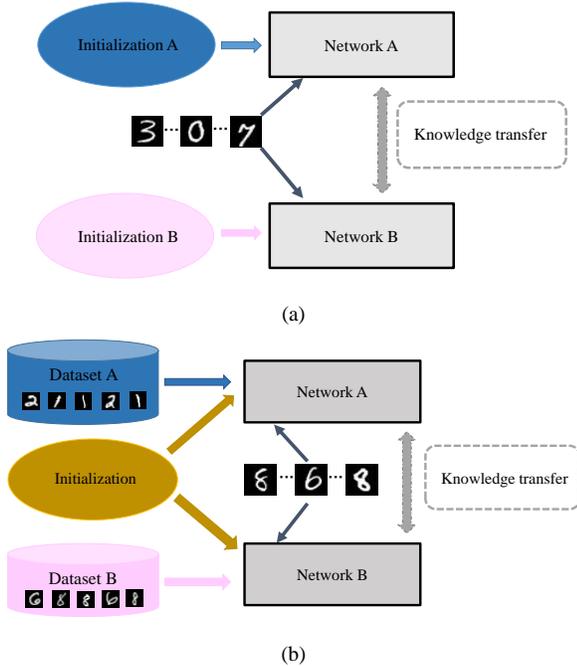

Fig. 3. (a) Schematic of DML [13]. Each network is initialized differently. (b) Schematic of MKT of the proposed Def-KT algorithm. All the networks are initialized similarly but trained on heterogeneous datasets before transferring knowledge in each round.

---

**Algorithm 2: FullAvg (Baseline)**

***Input:*** initial parameters $\mathbf{w}_0$

***Initialization:*** All clients are initialized with the same model with parameters $\mathbf{w}_0$

**For** $t = 0, ..., T$ **do**

  Randomly select a set $\mathcal{I}_t^A = \left\{ k_t^1, k_t^2, ..., k_t^Q \right\}$ of clients.

  Randomly select another set $\mathcal{I}_t^B = \left\{ k_t^{Q+1}, k_t^{Q+2}, ..., k_t^{2Q} \right\}$ of clients.

  **For** $j \in \{1, ..., Q\}$ **in parallel do**

$$\tilde{\mathbf{w}}_t^{k_t^j} \leftarrow \text{SGD}_{B_t, M} \left( \mathbf{w}_t^{k_t^j}, \mathcal{D}_{k_t^j} \right).$$

    The $k_t^j$-th client stores $\tilde{\mathbf{w}}_t^{k_t^j}$ as its local model and transmits it to the $k_t^{j+Q}$-th client.

  **end**

  **For** $j \in \{1, ..., Q\}$ **in parallel do**

    The $k_t^{j+Q}$-th client does:

$$\mathbf{w}_t^{k_t^{j+Q}} \leftarrow \frac{N_{k_t^j}}{N_{k_t^j} + N_{k_t^{j+Q}}} \tilde{\mathbf{w}}_t^{k_t^j} + \frac{N_{k_t^{j+Q}}}{N_{k_t^j} + N_{k_t^{j+Q}}} \mathbf{w}_t^{k_t^{j+Q}}.$$

    Store $\mathbf{w}_t^{k_t^{j+Q}}$ as the local model at the $k_t^{j+Q}$-th client.

  **end**

**end**

***Until:*** **convergence**

---

**Algorithm 3: Combo (Baseline)**

***Input:*** initial parameters $\mathbf{w}_0$

***Initialization:*** All clients are initialized with the same model with parameters $\mathbf{w}_0$

**For** $t = 0, ..., T$ **do**

  Randomly select a set $\mathcal{I}_t^A = \left\{ k_t^1, k_t^2, ..., k_t^Q \right\}$ of clients.

  Randomly select another set $\mathcal{I}_t^B = \left\{ k_t^{Q+1}, k_t^{Q+2}, ..., k_t^{2Q} \right\}$ of clients.

  **For** $j \in \{1, ..., Q\}$ **in parallel do**

$$\tilde{\mathbf{w}}_t^{k_t^j} \leftarrow \text{SGD}_{B_t, M} \left( \mathbf{w}_t^{k_t^j}, \mathcal{D}_{k_t^j} \right).$$

    The $k_t^j$-th client partitions $\tilde{\mathbf{w}}_t^{k_t^j}$ into two segments $\left\{ \tilde{\mathbf{w}}_{t,1}^{k_t^j}, \tilde{\mathbf{w}}_{t,2}^{k_t^j} \right\}$ and transmits $\tilde{\mathbf{w}}_{t,2}^{k_t^j}$ to the $k_t^{j+Q}$-th client.

  **end**

  **For** $j \in \{1, ..., Q\}$ **in parallel do**

    The $k_t^{j+Q}$-th client partitions $\mathbf{w}_t^{k_t^{j+Q}}$ into two segments $\left\{ \mathbf{w}_{t,1}^{k_t^{j+Q}}, \mathbf{w}_{t,2}^{k_t^{j+Q}} \right\}$ and transmits $\mathbf{w}_{t,1}^{k_t^{j+Q}}$ to the $k_t^j$-th client.

    The $k_t^{j+Q}$-th client does:

$$\mathbf{w}_{t,2}^{k_t^{j+Q}} \leftarrow \frac{N_{k_t^j}}{N_{k_t^j} + N_{k_t^{j+Q}}} \tilde{\mathbf{w}}_{t,2}^{k_t^j} + \frac{N_{k_t^{j+Q}}}{N_{k_t^j} + N_{k_t^{j+Q}}} \mathbf{w}_{t,2}^{k_t^{j+Q}}.$$

    Store $\left\{ \mathbf{w}_{t,1}^{k_t^{j+Q}}, \mathbf{w}_{t,2}^{k_t^{j+Q}} \right\}$ as the local model at the $k_t^{j+Q}$-th client.

    The $k_t^j$-th client does:

$$\tilde{\mathbf{w}}_{t,1}^{k_t^j} \leftarrow \frac{N_{k_t^j}}{N_{k_t^j} + N_{k_t^{j+Q}}} \tilde{\mathbf{w}}_{t,1}^{k_t^j} + \frac{N_{k_t^{j+Q}}}{N_{k_t^j} + N_{k_t^{j+Q}}} \mathbf{w}_{t,1}^{k_t^{j+Q}}.$$

    Store $\left\{ \tilde{\mathbf{w}}_{t,1}^{k_t^j}, \tilde{\mathbf{w}}_{t,2}^{k_t^j} \right\}$ as the local model at the $k_t^j$-th client.

  **end**

**end**

***Until:*** **convergence**

---

### B. Baseline Methods

Two popular baseline DFL methods based on model averaging are adopted for comparison. To make the baseline methods and the proposed method comparable, all the methods are implemented under the same assumptions as described in Section II, and the communication overhead per round of the three methods is set to be the same. The first baseline method is FullAvg given as Algorithm 2. In FullAvg, model fusion is implemented by averaging full sets of model parameters, as done in [7], [10] and [11], and the weights are determined by the sizes of the training datasets of different clients [2]. Note that in Algorithm 2 $N_{k_t^j}$ and $N_{k_t^{j+Q}}$ denote the number of data samples owned by the $k_t^j$-th client and the $k_t^{j+Q}$-th client, respectively. In the baseline algorithm Combo given as Algorithm 3, the models are fused by averaging model segments [36]. It can be seen that the overall transmission overhead between the $k_t^j$ -th client and the $k_t^{j+Q}$ -th client in the $t$-th round of algorithm Combo is a complete set of the model parameters, which is the same as that of FullAvg and Def-KT.

When the training data across different clients are drawn from the same statistical distribution and each client only performs a small number of SGD iterations to update its local model, model averaging in FullAvg and Combo performs well since the local updates of different clients are closely related to each other and model averaging does not suffer much from client-drift. However, practical FL systems rarely satisfy the above conditions due to the following reasons. First, datasets of different clients are generated in distinct manners and their training data follow different distributions [29]-[31]. Second, resulting from limited communication resources, the system designer is more inclined to perform multiple steps of SGD in each round to reduce the total number of rounds [2].

### C. Implementation Details

In our experiments, two types of neural networks are adopted, namely multi-layer perceptron (MLP) [2] and convolution neural network (CNN). The MLP has two hidden layers with 200 units, each of which uses ReLu activation. The MLP model contains 199,210 parameters. The CNN trained on the Fashion-MNIST dataset contains two convolution layers, a fully connected layer and a softmax layer, which has 29034 parameters. The CNN trained on the CIFAR-10 dataset contains three convolution layers, two fully connected layers and a softmax layer, which has 122570 parameters. The CNN trained on the CIFAR-100 dataset has three convolution layers, two fully connected layers and a softmax layer with 128420 parameters.

For all the experiments, with a total number of $K$ clients, the training set is divided into $K$ equal parts for all clients. Two settings are considered for the experiments, i.e., the homogeneous setting where the training data are independent and identically distributed (IID) across different clients and the heterogeneous setting where the training data are distributed across the clients in a non-IID manner. For the IID setting[4], the data samples in the training dataset are shuffled and randomly distributed to each of the $K$ clients. For the non-IID setting, most of the clients own data of only $\xi$ classes, and the value of $\xi$ will be specified in Section IV.D for each experiment. This non-IID setting is generated by arranging the training data by their labels, dividing them into $K\xi$ segments of equal size, and then assigning $\xi$ segments randomly to each client. It can be seen that a smaller value of $\xi$ indicate a higher degree of data heterogeneity across different clients. This data partitioning enables us to explore the robustness of our method to the data with highly heterogeneous distributions.

For each client, 80% of the local data are used as the private training data, and the remaining data is used as a local validation set which is used to test the local classification accuracy of the model on data samples drawn from its local distribution. Besides, the test set is used to test the global classification accuracy on data drawn from a joint distribution of all clients in the network. It is worth noting that, with non-IID training data, the global classification accuracy and the local classification accuracy are not necessarily the same. For example, with MNIST dataset and 10 clients in the network, each client obtains 4800 training samples and a local validation set of 1200 samples. And a global test set of 10k samples is used to test the global performance of the models of all the clients. In all experiments, the models at different clients are initialized with the same architecture and the same set of model parameters. For all the experiments, the participation rate is fixed at 20%, which means 20% of the clients participate in each round. This experimental setting complies with the common practice of partial participation used in FL problems in many practical scenarios [2]. During the MKT process in the proposed Def-KT algorithm, $E$=1 number of passes is made over the local dataset at the participating client who receives the model from another participating client in the current round. And when Def-KT is implemented, the learning rates $\eta_0$, $\eta_1$, and $\eta_2$ are set to the same value, which will be specified in Section IV.D under each setting.

### D. Experimental Results and Discussions

To compare the performances of the proposed method and the baseline methods with model averaging, we run four sets of experiments, whose results are shown as follows.

1) The MLP model is tested on the MNIST dataset in the IID setting and the non-IID setting ($\xi = 8$ and $\xi = 4$) with varying numbers of clients. In each round, the number of training passes in local updating is fixed at $M = 1$. An SGD optimizer with momentum = 0.5, learning rate = 0.01, and batchsize = 200 is applied. In Fig. 4(a)-(c), Fig. 4(e)-(g), and Fig. 4(i)-(k), for Def-KT, FullAvg and Combo, we depict the averaged global classification accuracy of the models of all the clients versus the number of rounds in the training stage with different numbers of clients. In Fig. 4(d), Fig. 4(h) and Fig. 4(l), the averaged local classification accuracy across all the clients for the three methods is reported when a certain number of rounds have been accomplished under the non-IID setting with varying numbers of clients. To prevent redundancy, the local classification accuracy under the IID setting is not presented in Fig. 4(d), Fig. 4(h) and Fig. 4(l), considering that the global classification accuracy and the local classification accuracy are equal to each other under the IID setting.

2) The CNN model is tested on the Fashion-MNIST dataset in the IID setting and the non-IID setting ($\xi = 8$ and $\xi = 4$), where the number of clients is fixed at $K = 10$. In each round, the number of training passes in the local updating is set as $M = 10$. An SGD optimizer with momentum = 0.5, learning rate = 0.001, and batchsize = 200 is applied for the IID setting and the non-IID setting ($\xi = 8$). An SGD optimizer with momentum = 0.5, learning rate = 0.005, and batchsize = 200 is applied for the non-IID setting ($\xi = 4$). In Fig. 5(a)-(c), for Def-KT, FullAvg and Combo, we plot the averaged global classification accuracy of the models at all clients versus the number of rounds. The values of the averaged local classification accuracy of all the methods are shown in Fig.

---

[4] In the rest of the paper, we use "IID" and "homogeneous" interchangeably. Likewise, we use "non-IID" and "heterogeneous" interchangeably.

5(d) under the two non-IID setting upon completing a certain number of rounds.

3) The CNN model is tested on the CIFAR-10 dataset in the IID setting and the non-IID setting ($\xi = 8$), where the number of clients is fixed at $K = 10$. In each round, the number of training passes in the local updating is fixed at $M = 1$ and an SGD optimizer with momentum $= 0.5$, learning rate $= 0.01$, and batchsize $= 200$ is applied. In Fig.

6(a) and Fig. 6(b), for Def-KT, FullAvg and Combo, we depict the averaged global classification accuracy of the models at all clients versus the number of rounds in the training process. In Fig. 6(c), we present the averaged local classification accuracy across all the clients for the three methods under the non-IID setting after implementing a certain number of rounds.

4) The CNN model is tested on the CIFAR-100 dataset in the

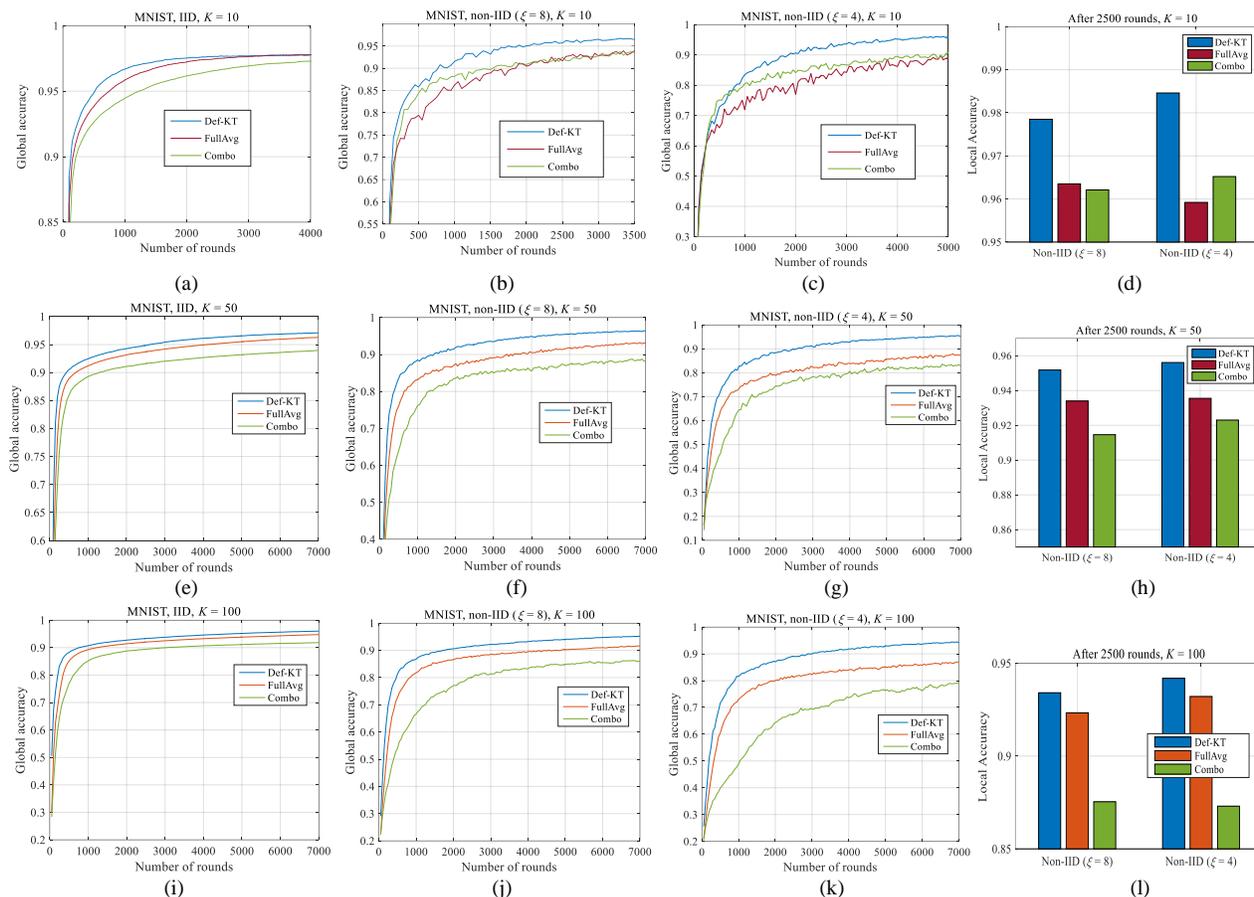

Fig. 4. Classification accuracy on the MNIST dataset using MLP, where different algorithms are performed. (a), (e) and (i) present the global classification accuracy versus the number of rounds under the IID case. (b),(f) and (j) present the global classification accuracy versus the number of rounds under the non-IID case where $\xi = 8$ . (c), (g) and (k) present the global classification accuracy versus the number of rounds under the non-IID case where $\xi = 4$ . (d), (h), and (l) present the local classification accuracy after 2500 rounds under two non-IID cases, where $\xi = 8$ and $\xi = 4$ . In (a)-(d), the number of clients is fixed at $K = 10$. In (e)-(h), the number of clients is fixed at $K = 50$. In (i)-(l), the number of clients is fixed at $K = 100$.

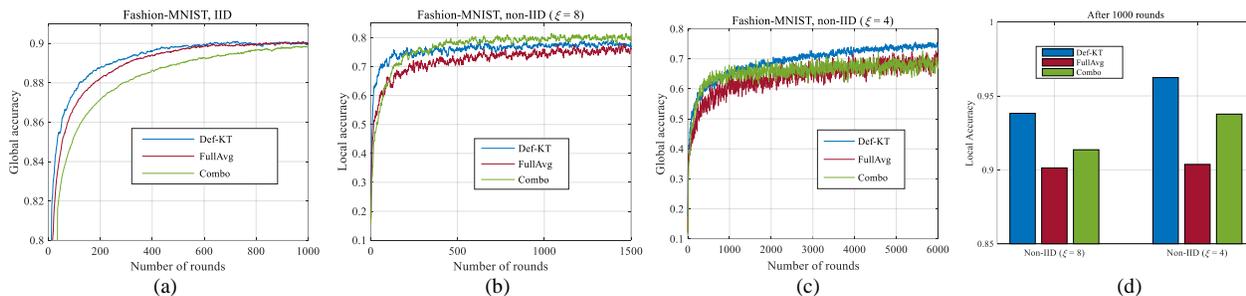

Fig. 5. Classification accuracy on the Fashion-MNIST dataset using CNN, where different algorithms are performed. (a) Global classification accuracy versus the number of rounds under the IID case. (b) Global classification accuracy versus the number of rounds under the non-IID case, where $\xi = 8$ . (c) Global classification accuracy versus the number of rounds under the non-IID case, where $\xi = 4$ . (d) Local classification accuracy after 1000 rounds under two non-IID cases, where $\xi = 8$ and $\xi = 4$ .

IID setting and the non-IID setting ($\xi = 50$), where the number of clients is fixed at $K = 10$. In each round, we fix the number of training passes in the local updating at $M = 1$ and an SGD optimizer with momentum = 0.5, learning rate = 0.01, and batchsize = 200 is used. In Fig. 7(a) and Fig. 7(b), for Def-KT, FullAvg and Combo, we plot the averaged global classification accuracy of the models at all clients versus the number of rounds in the training process. In Fig. 7(c), the averaged local classification accuracy of the three methods, which is attained after a certain number of rounds under the non-IID setting, is reported.

From Fig. 4-Fig. 7, it is obvious that the proposed Def-KT method outperforms the baseline methods based on the following facts.

1) From Fig. 4(a)-(c), Fig. 4(e)-(g), Fig. 4(i)-(k), Fig. 5(a)-(c), Fig. 6(a)-(b), and Fig. 7(a)-(b), the proposed Def-KT converges faster and attains a higher global classification accuracy after a fixed number of rounds in most cases compared with the baseline methods. This is because under data heterogeneity, model averaging induces client-drift which degrades the learning performance. By employing the proposed Def-KT method, the resulting model of MKT preserves the information from both models participating in the model fusion process. Moreover, it can be seen that the superiority of the proposed method is even more significant under the non-IID setting than that under the IID setting, which verifies the fact that Def-KT could make better use of data heterogeneity to enhance the generalization ability of the trained model while the baseline methods are not able to.

2) From Fig. 4(b)-(c), Fig. 4(f)-(g), Fig. 4(j)-(k), Fig. 5(b)-(c), Fig. 6(b) and Fig. 7(b), it can be seen that the global classification accuracy of the baseline methods oscillates more severely under the non-IID setting while that of the proposed method is more stable. Based on this observation, it is verified that, with heterogeneous data, transferring knowledge to fuse different models as done in Def-KT ensures the resulting model to gain improved generalization ability while averaging models directly induces much uncertainty into the performance of the resulting model. From this point of view, the proposed method is more reliable.

3) From Fig. 4(d), Fig. 4(h), Fig. 4(l), Fig. 5(d), Fig. 6(c) and Fig. 7(c), it can be observed that the local classification accuracy of the proposed method is significantly higher than that of the baseline methods. This is because during the MKT process of Def-KT, the resulting model successfully incorporates the knowledge of both models regarding their local datasets. On the contrary, model averaging at a local client leads to catastrophic forgetting of the previously learnt knowledge, since some of the weights in the network that are vital for the classification on its local dataset are incorrectly modified. Besides, the performance gap between the proposed method and the baseline methods is enlarged with increasing data heterogeneity, i.e., decreasing $\xi$, which demonstrates that the proposed method is more robust to highly heterogeneous data.

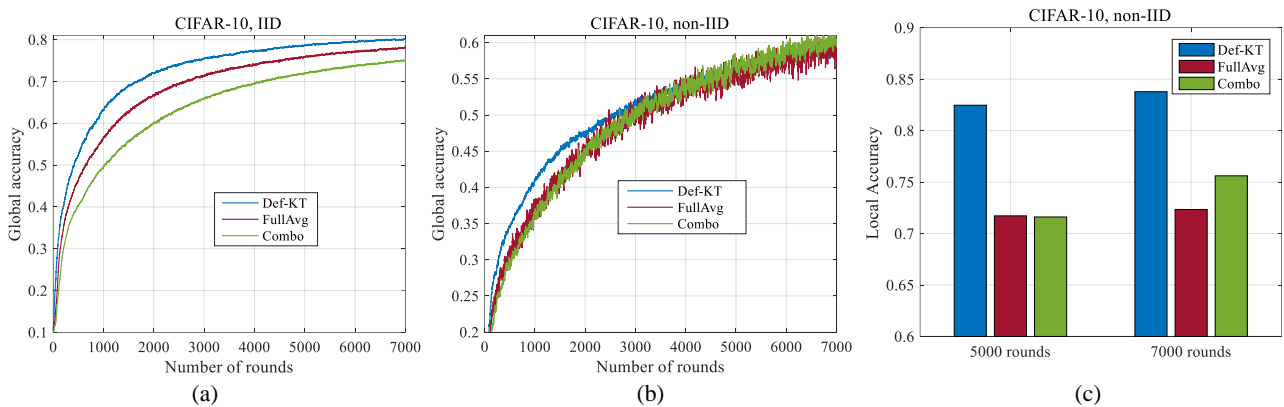

Fig. 6. Classification accuracy on the CIFAR-10 dataset using CNN, where different algorithms are performed. (a) Global classification accuracy versus the number of rounds under the IID case. (b) Global classification accuracy versus the number of rounds under the non-IID case where $\zeta = 8$. (c) Local classification accuracy after 5000 rounds and 7000 rounds under the non-IID case where $\xi = 8$.

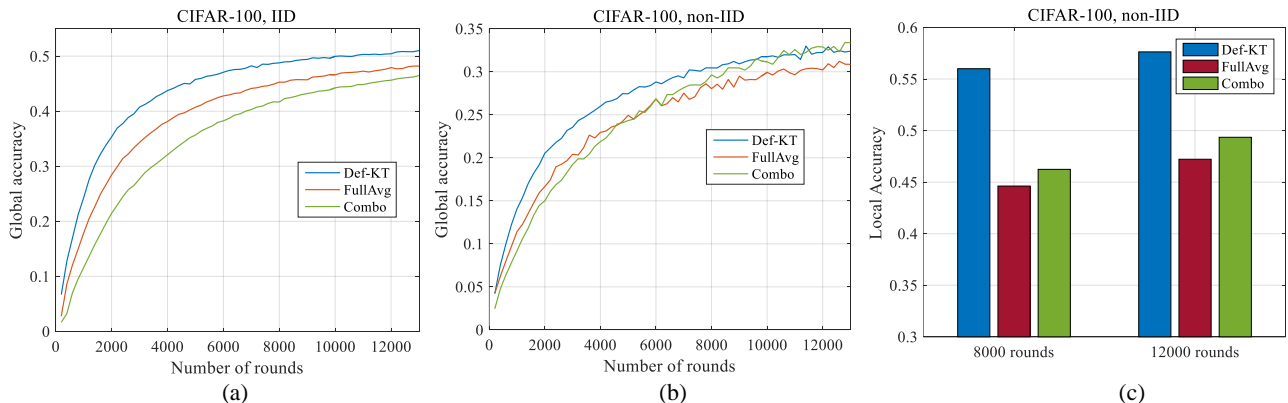

Fig. 7. Classification accuracy on the CIFAR-100 dataset using CNN, where different algorithms are performed. (a) Global classification accuracy versus the number of rounds under the IID case. (b) Global classification accuracy versus the number of rounds under the non-IID case where $\xi = 50$. (c) Local classification accuracy after 8000 rounds and 12000 rounds under the non-IID case where $\xi = 50$.

4) The experimental results in Fig. 4 demonstrate that the proposed Def-KT algorithm outperforms the baseline methods with both large and small number of clients in the network. From this perspective, Def-KT can be widely applied in many scenarios including horizontally FL to business (H2B) where there are only a handful of clients and horizontally FL to consumers (H2C) where more clients are involved in the learning process [52].

5) It can be seen from Fig. 5(b) that the proposed Def-KT algorithm attains almost the same performance as Combo. The reason can be explained as follows. On one hand, in each round of Combo, all the participating clients receive model segments from other clients. Based on that, model segments are fused at all the participating clients, all of which could indirectly gain some extra knowledge about the private datasets of other clients. In contrast, in Def-KT, only half of the participating clients transmit models and only the other half of the participating clients could gain knowledge from other clients by MKT. From this perspective, in each round, more clients could gain knowledge from other clients in Combo compared with Def-KT, which contributes to better learning performance of Combo. On the other hand, Combo averages model segments for model fusion, which induces client-drift and degrades its learning performance. In contrast, Def-KT adopts MKT to fuse different models, which avoids the negative impacts of client-drift. In Fig. 5(b), compared with Def-KT, the advantages and the disadvantages of Combo cancel out. Hence, Def-KT and Combo attain almost the same learning performance under that case. However, it is worth noting that all the other experimental results in this paper show that Def-KT significantly outperforms the two baseline methods, which demonstrates its superiority.

## V. CONCLUSION

In this paper, we considered the DFL problem in IoT systems, where a number of IoT clients train models for image classification tasks without the assistance of a central server. To avoid the degradation of the learning performance induced by client-drift under data heterogeneity, we proposed a novel Def-KT algorithm which incorporates the advantages of MKT into DFL schemes. We ran experiments on the MNIST, Fashion-MNIST, CIFAR-10 and CIFAR-100 datasets, which demonstrate the superiority of the proposed Def-KT algorithm over the baseline DFL methods based on model averaging, as the former attains higher classification accuracy and more stable learning performance. In the future, we plan to study more communication-efficient extensions of Def-KT and to conduct theoretical analysis for Def-KT.

## REFERENCES

[1] J. Konecný, H. B. McMahan, F. X. Yu, P. Richtárik, A. T. Suresh, and D. Bacon, "Federated learning: Strategies for improving communication efficiency," 2016. [Online]. Available: http://arxiv.org/abs/1610.05492.

[2] H. B. McMahan, D. Ramage, S. Hampson, and B. Aguera y Arcas, "Communication-efficient learning of deep networks from decentralized data," in *Proc. 20th Int. Conf. Artificial Intelligence and Statistics,* 2017, pp. 1273–1282.

[3] P. Kairouz *et al,* "Advances and open problems in federated learning," 2019. [Online]. Available: https://arxiv.org/abs/1912.04977.

[4] T. Li, A. K. Sahu, A. Talwalkar, and V. Smith, "Federated learning: Challenges, methods, and future directions," *IEEE Signal Process. Mag.,* vol. 37, no. 3, pp. 50–60, May 2020.

[5] S. Boyd, A. Ghosh, B. Prabhakar and D. Shah, "Randomized gossip algorithms," *IEEE Trans. Inf. Theory,* vol. 52, no. 6, pp. 2508-2530, Jun. 2006.

[6] M. Blot, D. Picard, M. Cord, and N. Thome, "Gossip training for deep learning," 2016. [Online]. Available: https://arxiv.org/abs/1611.09726.

[7] I. Hegedus, G. Danner, and M. Jelasity, "Gossip learning as a decentralized alternative to federated learning," in *Proc. Int. Conf. Distributed Applications and Interoperable Systems,* pages 74-90, 2019.

[8] A. Koloskova, S. U. Stich, and M. Jaggi, "Decentralized stochastic optimization and gossip algorithms with compressed communication," in *Proc. Int. Conf. Mach. Learn.,* 2019, pp. 3478–3487.

[9] A. Koloskova, N. Loizou, S. Boreiri, M. Jaggi, and S. U. Stich, "A unified theory of decentralized sgd with changing topology and local updates," 2020. [Online]. Available: https://arxiv.org/abs/2003.10422.

[10] A. G. Roy, S. Siddiqui, S. Polsterl, N. Navab, and C. Wachinger, "Braintorrent: a peer-to-peer environment for decentralized federated learning," 2019. [Online]. Available: https://arxiv.org/abs/1905.06731.

[11] S. Savazzi, M. Nicoli, V. Rampa, and S. Kianoush, "Federated learning with mutually cooperating devices: A consensus approach towards server-less model optimization," in *Proc. Int. Conf. Acoustics, Speech, and Signal Processing,* pages 3937-3941, 2020.

[12] S. Praneeth Karimireddy, S. Kale, M. Mohri, S. J. Reddi, S. U. Stich, and A. Theertha Suresh, "SCAFFOLD: Stochastic controlled averaging for federated learning," in *Proc. Int. Conf. Mach. Learn.,* 2020.

[13] Y. Zhang, T. Xiang, T. Hospedales, and H. Lu, "Deep mutual learning," in *Proc. IEEE Conf. Comput. Vis. Pattern Recognit.,* Jun. 2018, pp. 4320–4328.

[14] Y. LeCun, L. Bottou, Y. Bengio, and P. Haffner, "Gradient-based learning applied to document recognition," *Proc. IEEE,* vol. 86, no. 11, pp. 2278–2324, Nov. 1998.

[15] A. Krizhevsky, V. Nair, and G. Hinton. (2014). *The CIFAR-10 Dataset.* [Online]. Available: http://www.cs.toronto.edu/kriz/cifar.html.

[16] M. Xie, G. Long, T. Shen, T. Zhou, X. Wang, and J. Jiang, "Multi-Center Federated Learning," 2020. [Online]. Available: https://arxiv.org/abs/2005.01026.

[17] T. Li, A. K. Sahu, M. Zaheer, M. Sanjabi, A. Talwalkar, and V. Smith, "Federated optimization in heterogeneous networks," in *Proc. Conf. Machine Learning and Systems,* 2020.

[18] A. Romero, N. Ballas et al, "Fitnets: Hints for thin deep nets," in *Proc. Int. Conf. Learning Representations,* 2015.

[19] G. Hinton, O. Vinyals, and J. Dean, "Distilling the knowledge in a neural network," 2015. [Online]. Available: https://arxiv.org/abs/1503.02531.

[20] S. A. Rahman, H. Tout, H. Ould-Slimane, A. Mourad, C. Talhi and M. Guizani, "A Survey on Federated Learning: The Journey from Centralized to Distributed On-Site Learning and Beyond," *IEEE Internet Things J.,* early access, 2020.

[21] J. Pang, Y. Huang, Z. Xie, Q. Han and Z. Cai, "Realizing the Heterogeneity: A Self-Organized Federated Learning Framework for IoT," *IEEE Internet Things J.,* early access, 2020.

[22] I. Mohammed et al., "Budgeted Online Selection of Candidate IoT Clients to Participate in Federated Learning," *IEEE Internet Things J.,* early access, 2020.

[23] R. Saha, S. Misra and P. K. Deb, "FogFL: Fog Assisted Federated Learning for Resource-Constrained IoT Devices," *IEEE Internet Things J.,* early access, 2020.

[24] Y. Zhao, M. Li, L. Lai, N. Suda, D. Civin, and V. Chandra, "Federated learning with non-IID data," 2018. [Online]. Available:

https://arxiv.org/abs/1806.00582.

[25] H. Xiao, K. Rasul, and R. Vollgraf, "Fashion-MNIST: a Novel Image Dataset for Benchmarking Machine Learning Algorithms," 2017. [Online]. Available: https://arxiv.org/abs/1708.07747.

[26] K. Bonawitz et al., "Towards federated learning at scale: System design," in *Proc. 2nd SysML Conference*, 2019.

[27] A. D. G. Dimakis, A. D. Sarwate and M. J. Wainwright, "Geographic Gossip: Efficient Averaging for Sensor Networks," *IEEE Trans. Signal Process.*, vol. 56, no. 3, pp. 1205-1216, Mar. 2008.

[28] A. Khosravi and Y. S. Kavian, "Broadcast Gossip Ratio Consensus: Asynchronous Distributed Averaging in Strongly Connected Networks," *IEEE Trans. Signal Process.*, vol. 65, no. 1, pp. 119-129, Jan. 2017.

[29] F. Sattler, S. Wiedemann, K. -R. Müller and W. Samek, "Robust and Communication-Efficient Federated Learning From Non-i.i.d. Data," *IEEE Trans. Neural Netw. Learn. Syst.*, vol. 31, no.9, pp. 3400-3413, 2020.

[30] D. Li and J. Wang, "FedMD: Heterogenous federated learning via model distillation," in *Proc. 33rd Conf. Neural Information Processing Systems (NeurIPS)*, 2019.

[31] N. Shoham et al., "Overcoming forgetting in federated learning on non-iid data," in *NeurIPS Workshop on Federated Learning for Data Privacy and Confidentiality*, 2019.

[32] R. French, "Catastrophic forgetting in connectionist networks," *Trends in cognitive sciences*, vol. 3, pp. 128–135, 1999.

[33] Z. Li and D. Hoiem, "Learning without forgetting," *IEEE Trans. Pattern Anal. Mach. Intell.*, vol. 40, no. 12, pp. 2935–2947, Dec. 2018.

[34] J. Kirkpatrick et al., "Overcoming catastrophic forgetting in neural networks," *Proceedings of the National Academy of Sciences of the United States of America (PNAS)*, vol. 114, no. 13, pp. 3521–3526, 2017.

[35] S. Feng et al, "Collaborative Group Learning," 2020. [Online]. Available: https://arxiv.org/abs/2009.07712.

[36] C. Hu, J. Jiang and Z. Wang, "Decentralized Federated Learning: A Segmented Gossip Approach," *International Workshop on Federated Learning for User Privacy and Data Confidentiality in Conjunction with IJCAI*, 2019.

[37] P. Vanhaesebrouck, A. Bellet, and M. Tommasi, "Decentralized collaborative learning of personalized models over networks," in *Proc. Int. Conf. Artificial Intelligence and Statistics (AISTATS)*, 2017.

[38] T. Nishio and R. Yonetani, "Client Selection for Federated Learning with Heterogeneous Resources in Mobile Edge," in *Proc. IEEE Int. Conf. Communications (ICC)*, Shanghai, China, 2019, pp. 1-7.

[39] L. Lou, Q. Li, Z. Zhang, R. Yang and W. He, "An IoT-Driven Vehicle Detection Method Based on Multisource Data Fusion Technology for Smart Parking Management System," *IEEE Internet Things J.*, vol. 7, no. 11, pp. 11020-11029, Nov. 2020.

[40] Ş. Kolozali et al., "Observing the Pulse of a City: A Smart City Framework for Real-Time Discovery, Federation, and Aggregation of Data Streams," *IEEE Internet Things J.*, vol. 6, no. 2, pp. 2651-2668, Apr. 2019.

[41] M. Mohammadi, A. Al-Fuqaha, M. Guizani and J. Oh, "Semisupervised Deep Reinforcement Learning in Support of IoT and Smart City Services," *IEEE Internet Things J.*, vol. 5, no. 2, pp. 624-635, Apr. 2018.

[42] W. Xu et al., "The Design, Implementation, and Deployment of a Smart Lighting System for Smart Buildings," *IEEE Internet Things J.*, vol. 6, no. 4, pp. 7266-7281, Aug. 2019.

[43] T. He, C. Shen, Z. Tian, D. Gong, C. Sun, and Y. Yan, "Knowledge adaptation for efficient semantic segmentation," in *Proc. IEEE Conf. Comput. Vis. Pattern Recognit.*, pp. 578-587, 2019.

[44] Y. Liu, K. Chen, C. Liu, Z. Qin, Z. Luo, and J. Wang, "Structured knowledge distillation for semantic segmentation," in *Proc. IEEE Conf. Comput. Vis. Pattern Recognit.*, pp. 2599-2608, 2019.

[45] T. Wang, L. Yuan, X. Zhang, and J. Feng, "Distilling object detectors with fine-grained feature imitation," in *Proc. IEEE Conf. Comput. Vis. Pattern Recognit.*, pp. 4928-4937, 2019.

[46] F. Zhang, X. Zhu, and M. Ye, "Fast human pose estimation," in *Proc. IEEE Conf. Comput. Vis. Pattern Recognit.*, pp. 3512-3521, 2019.

[47] Y. Liao, X. Shen and H. Rao, "Analytic Sensor Rules for Optimal Distributed Decision Given K-Out-of-L Fusion Rule Under Monte Carlo Approximation," *IEEE Trans. Autom. Control*, vol. 65, no. 12, pp. 5488-5495, Dec. 2020

[48] J. Xie, J. Fang, C. Liu and L. Yang, "Unsupervised Deep Spectrum Sensing: A Variational Auto-Encoder Based Approach," *IEEE Trans. Veh. Technol.*, vol. 69, no. 5, pp. 5307-5319, May 2020.

[49] J. Mills, J. Hu and G. Min, "Communication-Efficient Federated Learning for Wireless Edge Intelligence in IoT," *IEEE Internet of Things*

*J.*, vol. 7, no. 7, pp. 5986-5994, Jul. 2020.

[50] F. Samie, L. Bauer and J. Henkel, "From Cloud Down to Things: An Overview of Machine Learning in Internet of Things," *IEEE Internet of Things J.*, vol. 6, no. 3, pp. 4921-4934, Jun. 2019.

[51] A. Krizhevsky and G. Hinton, "Learning multiple layers of features from tiny images," *Technical report*, University of Toronto, 2009.

[52] L. Lyu, H. Yu, and Q. Yang, "Threats to federated learning: A survey," 2020, arXiv:*2003.02133*. [Online]. Available: http://arxiv.org/abs/2003.02133.

[53] J. Wang et al, "Tackling the Objective Inconsistency Problem in Heterogeneous Federated Learning," in *Proc. 34th Conf. Neural Information Processing Systems (NeurIPS)*, 2020.

[54] Y. Deng et al, "Distributionally Robust Federated Averaging," in *Proc. 34th Conf. Neural Information Processing Systems (NeurIPS)*, 2020.